\documentclass[letterpaper]{article} %
\usepackage{aaai20}  %

\usepackage{booktabs} %
\usepackage{xspace}
\usepackage{subfig}

\usepackage{graphicx}
\usepackage{amsfonts} %
\usepackage{xcolor}
\usepackage{CJKutf8}
\usepackage{multirow}
\usepackage{multicol}
\usepackage{float}
\usepackage{hyphenat}

\usepackage{times}  %
\usepackage{helvet} %
\usepackage{courier}  %
\usepackage[hyphens]{url}  %
\usepackage{amsmath}
\usepackage{comment}
\usepackage{array}
\usepackage{color}
\usepackage{mathtools}
\usepackage{dsfont}
\usepackage{wasysym}

\title{Multi-step Joint-Modality Attention Network for Scene-Aware Dialogue System}

\author{
\begin{tabular}[c]{@{}c@{}}
    Yun-Wei Chu\textsuperscript{1}~~~
    Kuan-Yen Lin\textsuperscript{2}~~~
    Chao-Chun Hsu\textsuperscript{3}~~~
    Lun-Wei Ku\textsuperscript{1}~~~
    \end{tabular}\\
    \textsuperscript{1}{Academia Sinica},~~~
    \textsuperscript{2}{Cornell Tech},~~~
    \textsuperscript{3}{University of Colorado Boulder},\\
    yunweichu@iis.sinica.edu.tw, kl924@cornell.edu, chao-chun.hsu@colorado.edu, lwku@iis.sinica.edu.tw
}    

\begin{document}

\maketitle
\begin{abstract}

Understanding dynamic scenes and dialogue contexts in order to converse with users has been challenging for multimodal dialogue systems. The 8-th Dialog System Technology Challenge (DSTC8) \cite{DSTC8} proposed an Audio Visual Scene-Aware Dialog (AVSD) task \cite{hori2018end}, which contains multiple modalities including audio, vision, and language, to evaluate how dialogue systems understand different modalities and response to users. In this paper, we proposed a multi-step joint-modality attention network (JMAN) based on recurrent neural network (RNN) to reason on videos. Our model performs a multi-step attention mechanism and jointly considers both visual and textual representations in each reasoning process to better integrate information from the two different modalities. Compared to the baseline released by AVSD organizers, our model achieves a relative 12.1\% and 22.4\% improvement over the baseline on ROUGE-L score and CIDEr score.

\end{abstract}

\maketitle
\section{Introduction}

Understanding visual information along with natural language have been a recent surge of interest in visual-textual applications, such as image-based visual question answering (VQA) and image-based visual dialogue question answering. In contrast to image-based VQA, where the model aims to response the answer of a single question for the given image, image-based visual dialogue question answering was introduced to hold a meaningful dialogue with users about the given image. However, because a single image is far less than enough to represent the details of an event, videos are commonly used to record what has happened. Therefore, reasoning based on a video is also worth exploring.

Because of the relatively large complex feature space, video-language tasks are more challenging than traditional image-language tasks. To be more specific, processing videos involves diverse objects, action flows, audio that are not issues for image processing. Similar to image-based VQA, video question answering answers a single question based on a given video. Video dialogue question answering, by contrast, reasons the dialogue as well as the sequential question-answer pairs it contains in order to answer the current question for the given video.

\begin{figure}[t]
\begin{center}
\includegraphics[scale=0.55, trim={0 0 0 0}]{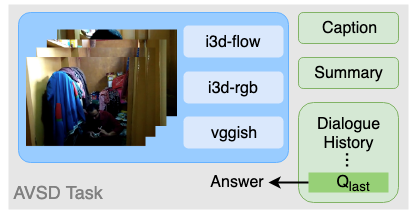}
\end{center}
   \caption{An illustration of DSTC8 AVSD task. The goal of the task is generating accurate answer based on multiple modalities.}
\label{fig:video_dialogue_example}
\end{figure}

The 8-th Dialog System Technology Challenge (DSTC8) Audio Visual Scene-Aware Dialogue (AVSD) task proposed a dataset to test the capability of dialogue responses with multiple modalities. A brief illustration of AVSD task is shown in Figure~\ref{fig:video_dialogue_example}. The task provides pre-extracted features using I3D \cite{DBLP:journals/corr/CarreiraZ17} and Vggish \cite{DBLP:journals/corr/HersheyCEGJMPPS16} models for the video. Moreover, a video caption, a video summary, and a dialogue history with question-answer pairs are introduced as textual information. Table \ref{tb:AVSD_example} shows an example of dialogue history, caption, summary from the AVSD training set. The purpose of this task is answering the question based on given multiple modalities.

\begin{table*}[h]
\begin{center}

\scalebox{0.8}{
\begin{tabular}{lll} 
\hline
Video Caption & \multicolumn{2}{l}{person a is in a pantry fixing a faulty camera . person a puts down the camera onto a pillow and closes the door .} \\

 \hline
 Video Summary & \multicolumn{2}{l}{a man is sitting in a closet fiddling with a camera . he puts the camera on the floor , gets up and walks out of the closet .}\\
  \hline
  \hline
   & \multicolumn{1}{c}{Question} & \multicolumn{1}{c}{Answer} \\ \cline{2-3}
 
    \multirow{9}{*}{Dialogue History} 
    & how many people are in this video ? & i can only see one in the video . \\ \cline{2-3} 
    & what is the setting of the video ? & a man is sitting in a closet fixing something .\\ \cline{2-3} 
    & can you tell what he is fixing ? & i think it is a camera .\\ \cline{2-3} 
    & does he sit in the closet the whole time ? & no , he gets out of the closet eventually .\\ \cline{2-3} 
    & where does he go to ? & outside of the closet but i do not know in which room he is afterwards .\\ \cline{2-3} 
    & is there audio ? & i do not hear anything .\\ \cline{2-3} 
    & does he take the camera with him when he exits the closet ? & no . the camera remains on the floor of the closet .\\ \cline{2-3} 
    & can you tell if he succeeds in fixing the camera ? & to be honest , i am not sure .\\ \cline{2-3} 
    & how does the video end ? & he is standing in the room doing nothing .
 
    \\ \hline

\end{tabular}}
\end{center}
\caption{A sample of a caption, a summary and a dialogue history of the video from DSTC8 AVSD dataset}
\label{tb:AVSD_example}
\end{table*}

In our work, we implement attention mechanisms \cite{bahdanau2014neural,DBLP:journals/corr/XuBKCCSZB15}, which have been proven useful for vision-language tasks, to focus on a rather important part in sources and to generate accurate answers on AVSD dataset. In order to increase the performance when the answer lies in a specific region of the video, our model performs multiple reasoning steps based on recurrent neural network (RNN) to find important representation. Moreover, to improve the understanding when the number of feature types increases, we proposed a joint-modality attention network (JMAN) to jointly learn attention from different features of the video. In conclusion, the results show that our model achieves a relative 12.1\% and 22.4\% improvement over the baseline on ROUGE-L score and CIDEr score.

\section{Related Work}
The Audio Visual Scene-Aware Dialog (AVSD) task aims at answering a free-form question based on the given video and texts. Therefore, we briefly review the vision-based question answering work in the following section.

\subsection{Visual Question Answering}
Given a natural-language question that targets visual features, image-based visual question answering (VQA) is to  provide an accurate answer relevant to the question. Because systems need to identify the most relevant region in the visual features based on question semantics, attention mechanisms show a powerful ability to focus on salient regions. A large amount of work \cite{DBLP:journals/corr/XuS15a,DBLP:journals/corr/YangHGDS15,DBLP:journals/corr/LuYBP16,DBLP:journals/corr/AndersonHBTJGZ17,DBLP:journals/corr/abs-1708-01471,DBLP:journals/corr/abs-1805-07932} demonstrate significant results on image-based VQA by attention mechanisms.

Das et. al first introduced the visual dialogue dataset (VisDial) \cite{visdial} which contains images from the COCO dataset \cite{DBLP:journals/corr/LinMBHPRDZ14} and one dialogue. Visual dialogue question answering aims to increase the ability of human-machine interaction by taking previous conversations into account. To capture important regions of visual features, attention mechanisms also play a role in visual dialogue question answering task, such as performing dynamic attention combination \cite{DBLP:journals/corr/abs-1709-07992}, recursively increasing visual co-reference resolution \cite{DBLP:journals/corr/abs-1812-02664}, implementing multiple reasoning steps on image and dialogue \cite{DBLP:journals/corr/abs-1902-00579}, and employing a multi-head attention mechanism \cite{DBLP:journals/corr/abs-1902-09368}.

\subsection{Video Question Answering}

Moving from image-based VQA to video question answering requires models to analyze relevant objects in the frames and keep track of temporal events. Much research \cite{DBLP:journals/corr/TapaswiZSTUF15,DBLP:journals/corr/abs-1809-01696,DBLP:journals/corr/JangSYKK17} provides video datasets form movies or TV series for systems to output an accurate answer given a set of potential answers. To answer question for videos, many approaches \cite{DBLP:journals/corr/YeZLCXZ17,DBLP:journals/corr/abs-1806-01873,DBLP:journals/corr/KimHCZ17,DBLP:journals/corr/abs-1709-09345} also utilize complicated attention mechanisms that focus on the most important part of videos.

In contrast to video question answering, video dialogue question answering task needs to understand dynamic scenes and previous conversations. The limited availability of such data makes this task more challenging. Recently, Hori et al. proposed an audio visual scene-aware dialog (AVSD) track in the 8-th Dialog System Technology Challenge (DSTC8). The AVSD dataset provides multimodal features, including vision, audio, and dialogue history, for videos. Table \ref{tb:various_dataset} shows the difference between AVSD dataset and other video datasets.
Instead of answering single question of the video, AVSD dataset takes historical question-answer pairs into account in order to generate a more conversation-like answer. Moreover, most of the video dataset select an answer from multiple choice, the AVSD dataset provides a free-form answer that makes the task more diffcult.

\begin{table}
\begin{center}
\scalebox{0.83}{
\begin{tabular}{lcc}
\hline
 Video QA Dataset  & Textual Format & Answer Form   \\
 \hline\hline
 
 MovieQA \cite{DBLP:journals/corr/TapaswiZSTUF15}  & QA  & multiple choice  \\
 TVQA \cite{DBLP:journals/corr/abs-1809-01696}  & QA  & multiple choice  \\
 TGIF\_QA \cite{DBLP:journals/corr/JangSYKK17}  & QA  & multiple choice  \\
 AVSD \cite{hori2018end}  & QA-dialogue  & free form  \\

 \hline
\end{tabular}}
\end{center}
\caption{The summary of several video question answering datasets.}
\label{tb:various_dataset}
\end{table}

\section{Proposed Approach}

\begin{figure*}[t!]
  \centering
  \subfloat[Overall Model Flowchart]{\raisebox{-0ex}{ \includegraphics[height=0.2\textheight,width=0.5\textwidth]{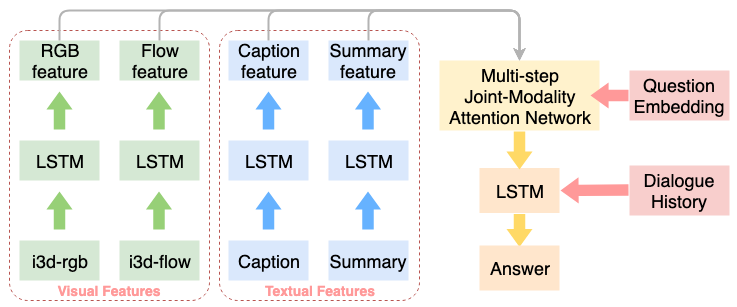}}\label{fig:f1}}
  \hfill
  \subfloat[Multi-Step Joint-Modality Attention Network]{\includegraphics[height=0.20\textheight,width=0.46\textwidth]{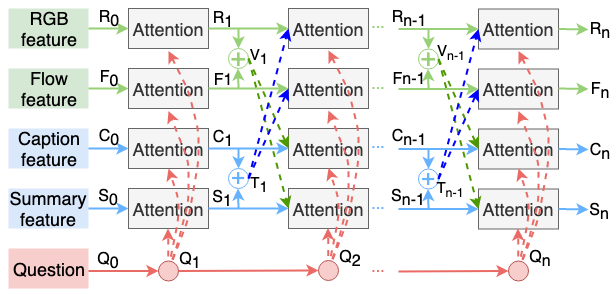}\label{fig:f2}}
  \caption{ In (a), every features are encoded by corresponding LSTM-based encoders. The proposed multi-step joint-modality attention network (JMAN) then learns attention from both visual and textual features. By considering previous conversation, our model then generate an answer by a LSTM-based decoder. A detailed illustration of proposed JMAN is shown in (b). Our proposed JMAN considers joint-attended features (\(V_n \) and \(T_n \)) in each reasoning step to increase video understanding from both visual and textual modalities.
  }
  \label{FIG1}
  \vspace{-1em}
\end{figure*}

Figure \ref{FIG1}(a) shows an overview of the proposed method. First, the model uses LSTM-based encoders to encode the visual features and textual features provided by AVSD organizers. We did not select audio feature proposed by organizers and we will explain in the Experiments section. Our proposed joint-modality attention network (JMAN) then attends the question with both visual features and textual representations. With the increasing recurrent reasoning steps of JMAN, the model learns the important visual regions and salient textual parts that correspond to the query. Finally, by jointly considering both visual and textual features, a LSTM-based decoder then generates an open-ended answer that best fits the given question, video, and context.

\subsection{Feature Extraction}
For visual features of videos, the AVSD organizers provide i3d-rgb and i3d-flow, which are extracted from the  “Mixed\_5c" layers of two-stream inflated 3D ConvNets \cite{DBLP:journals/corr/CarreiraZ17}. The visual features contain RGB information in each frame and optical flow information between frames. We use LSTM-based encoder with 2048 dimension to encode these two features. The encoded RGB feature and optical flow feature are denoted as \(R_0\) and \(F_0\).

Though we did not take audio feature to construct our final model, we still conduct experiments to evaluate the effectiveness of each features. In order to test the usefulness of the audio feature, which is extracted from Vggish model \cite{DBLP:journals/corr/HersheyCEGJMPPS16}, we also utilize LSTM-based encoder with 128 dimension to encode audio feature. The encoded audio feature represents as \(A_0\) for experimental purpose.

For the question, the caption, the summary, and the dialogue history of the AVSD dataset, we transferred each text into a vector using GloVE \cite{pennington2014glove}. All the textual vectors then encoded by 128 dimensional LSTM-based encoders to output encoded features of caption, summary, question, and dialogue history, and they are denoted as \(C_0\), \(S_0\), \(Q_0\), and \(D\) respectively.

\subsection{Multi-step Joint-Modality Attention Network}
An overview of the proposed multi-step joint-modality attention network (JMAN) is given in Figure \ref{FIG1}(b). The framework is based on a recurrent neural network (RNN), where the hidden state \(Q_n\) indicates the current question representation and the lower index \(n\) is the number of reasoning steps. After \(n\)-step attention mechanism, the attended RGB feature and the attended optical flow feature are represented as \(R_n\) and \(F_n\). Likewise, \(C_n\) and \(S_n\) are the attended caption feature and the attended summary feature. Specifically, we sum \(R_n\) and \(F_n\) as joint-attended visual feature \(V_n\) after reasoning step \(n\)=1; likewise, \(C_n\) and \(S_n\) are aggregated as the joint-attended textual feature \(T_n\). From the second reason step (\(n\) = 2), the joint-attended features will deliver to different modality to enhance both domains understanding. Take the second reasoning step (\(n\) = 2) as example, the joint-attended textual feature \(T_1\) will deliver to visual modality to attend the second question state \(Q_2\) together with \(R_1\) and \(F_1\). In contrast to attending to a single-domain modality with the query, we find that jointly attending different domain modality enhances the performance of video understanding. Moreover, proposed JMAN can focus on the salient region of both visual and textual features when the number of reasoning step increases.

\subsubsection{Self-Attended Question}
We applied self-attention to the current question representation \(Q_n\) which is the hidden state of proposed RNN-based JMAN.
\begin{equation} 
    \alpha_{Q} = \textup{softmax}(p_{Q} \cdot \tanh(\omega_{Q}Q_{n-1}^T)),
\end{equation}

\begin{equation} 
    Q_{n} = \alpha_{Q} \cdot Q_{n-1},
\end{equation}
where the attention score of question is \(\alpha_Q\) and the parameter matrices are \(p_Q\) and \(\omega_Q\).

\subsubsection{Attending Question and Previous Joint-Attended Features to Different Modalities}

The model updates attended RGB feature \(R_n\) and attended optical flow feature \(F_n\) by their previous state (\(R_{n-1}\) and \(F_{n-1}\)) and the current query \(Q_{n}\). The joint-attended textual feature \(T_{n}\) will also pass to the attention mechanism after the first reasoning step. In the following equations, we use index \(x \in \{R, F\}\) represents visual components (RGB and optical flow).

\begin{equation} 
    \alpha_{x} = \textup{softmax}(p_{x} \cdot \tanh(\omega_{x}x_{n-1}^T + \omega_{Q}'Q_{n}^T+\omega_{T}T_{n-1}^T)),
\end{equation}

\begin{equation} 
    x_{n} = \alpha_{x} \cdot x_{n-1},
\end{equation}
where \(\alpha_{x}\) is the attention score of the visual components, and the parameter matrices are \(p_x\),  \(\omega_x\), \(\omega_Q'\), and \(\omega_T\). The joint-attended textual feature \(T_n\) is delivered from the textual modality. After the first reasoning step, the model begins to aggregate \(R_n\) and \(F_n\) as joint-attended visual feature \(V_n\), which is delivered to the textual modality.

Similar to the attention mechanism for visual modality, the model updates attended caption feature \(C_n\) and attended summary  feature \(S_n\) by their previous state (\(C_{n-1}\) and \(S_{n-1}\)) and the current query \(Q_{n}\). The joint-attended visual feature \(V_n\) transfers into textual modality in order to use the salient visual information to discover important textual information. We use index \(y \in \{C, S\}\) represents textual components (caption and summary).
\begin{equation} 
    \alpha_{y} = \textup{softmax}(p_{y} \cdot \tanh(\omega_{y}y_{n-1}^T + \omega_{Q}''Q_{n}^T+\omega_{V}V_{n-1}^T)),
\end{equation}

\begin{equation} 
    y_{n} = \alpha_{y} \cdot y_{n-1},
\end{equation}
where \(\alpha_{y}\) is the attention score of the textual components, and the parameter matrices are \(p_y\),  \(\omega_y\), \(\omega_Q''\), and \(\omega_V\). The joint-attended visual feature \(V_n\) is delivered from the visual modality. The system begins to sum \(C_n\) and \(S_n\) as joint-attended textual feature \(T_n\) after reasoning step \(n\) = 1, and \(T_n\)  will pass to the visual modality as additional information.

\subsection{Answer Decoder}
The system concatenates all attended features \(R_n\), \(F_n\), \(C_n\), and \(S_n\) as the context vector \(z_n\). The question representation is updated based on context vector via an RNN with Gate Recurrent Unit (GRU) \cite{cho-etal-2014-learning}:
\begin{equation} 
   Q_{n+1} = GRU (Q_n, z_n).
\end{equation}
A generative LSTM-based decoder is used to decode the context vector \(z_n\). 
Each question-answer pair in dialogue history will also be used to generate the answer \(a = ( a_{1},  a_{2},...,  a_{L})\), where \(L\) is the number of word, and \(a_{\ell} \in \Gamma_{\ell} = \{1, 2,..., |\Gamma_{\ell}|\}\) represents the a vocabulary of possible words \(\Gamma_{\ell}\). 
By considering the context vector \(z_n\) and dialogue history \(D\), an FC-layer with dropout and softmax is used after the decoder to compute the conditional probability \( p(a_{\ell}| D_, a_{\ell-1}, h_{\ell-1} )\) for possible word \(a_{\ell}\), where the initial hidden state \(h_0\) is \(z_n\).

\section{Experiments and Results}

\subsection{Experimental Materials and Setup}
The organizers of DSTC8-AVSD track provide DSTC7-AVSD dataset for model constructing. From Charades video dataset \cite{DBLP:journals/corr/SigurdssonVWFLG16}, the AVSD dataset proposes for each corresponding video a dialog with 10 question-answer pairs, visual features generated by the I3D model \cite{DBLP:journals/corr/CarreiraZ17}, and audio feature produced by Vggish model \cite{DBLP:journals/corr/HersheyCEGJMPPS16}. The dialogue was generated via a discussion between two Amazon Mechanical Turk workers about the events observed in the video. Table \ref{tb:DSTC_data} summarizes the data distribution of the AVSD dataset.

\begin{table}
\begin{center}
\scalebox{0.9}{
\begin{tabular}{cccc}
\hline
   & Training & Validation  & Test \\
 \hline\hline
 \# of Dialogs & 7,659 & 1,787  & 1,710  \\
 \# of Turns & 153,180 & 35,740  & 13,490  \\
 \# of Words & 1,450,754 & 339,006  & 110,252   \\

\hline
\end{tabular}}
\end{center}
\caption{The data distribution of AVSD dataset.}
\label{tb:DSTC_data}
\end{table}

For our submitted system, we only select the visual features and textual features proposed by AVSD dataset to build our model. The dimensions of textual and visual features are set to 128 and 2048, and we use Adam optimizer \cite{kingma2014adam} with a learning rate of 0.001 in the training process. The batch size and a dropout rate \cite{Srivastava:2014:DSW:2627435.2670313} of proposed model is set to 32 and 0.2. Cross-entropy loss between the prediction and target are used to optimize the hyperparameter.

\subsection{Features Effectiveness}

To evaluate the influence of multimodal features on the AVSD task, we began by inputting dialogue history feature and then adding other mono-type features. We first considered the question and dialogue history, and the result of this simplest model (JMAN(DH)) is shown in the second part of Table \ref{tb:mono}. Without any attention mechanism on the features, JMAN(DH) ouputs answers based on dialogue history and performs poor than all other models with additional mono-type feature. This result is reasonable because of the insufficient information of video-related features. In order to further analyze the effectiveness of each feature, we add mono-type features on JMAN(DH) and set the reasoning step to 1. Therefore,  the attention algorithms are rewritten as : 

\begin{equation} 
    \alpha_{M} = \textup{softmax}(p_{M} \cdot \tanh(\omega_{M}M_{0}^T + \widehat{\omega}_{Q}Q_{1}^T),
\end{equation}

\begin{equation} 
    M_{1} = \alpha_{M} \cdot M_{0},
\end{equation}
where \(M\in \{A, R, F, C, S\} \) represent the feature components (audio, RGB, optical flow, caption, summary), and the parameter matrices are \(p_M\), \(\omega_{M}\), and \(\widehat{\omega}_{Q}\). As shown in the second part of Table \ref{tb:mono}, all models with additional mono-type feature outperform the simplest model JMAN(DH). This result shows the effectiveness of single-step attention mechanism on additional mono-type feature. Moreover, as it is likely that the question concerns what happens in the video, all models considering video-related components performs better than the simplest model.

\begin{table*}
\begin{center}
\small
\scalebox{1.1}{
\begin{tabular}{cccccccc}
\hline
   & BLEU-1  & BLEU-2  & BLEU-3  & BLEU-4 & METEOR & ROUGE-L & CIDEr   \\
 \hline
  \multicolumn{1}{l}{Baseline ~\cite{hori2018end}}&  0.621&	0.480&	0.379&	0.305&	0.217&	0.481&	0.733      \\ 
 \hline\hline
 \multicolumn{8}{c}{single reasoning step (\(n\) = 1)}\\
 \hline
   \multicolumn{1}{l}{JMAN (DH)} & 0.623 & 0.478 & 0.374 & 0.295 & 0.220 & 0.492 & 0.775  \\
   \multicolumn{1}{l}{JMAN (DH, C)} & 0.632 & 0.486 & 0.381 & 0.303 & 0.238 & 0.518 & 0.896\\
   \multicolumn{1}{l}{JMAN (DH, S)} & 0.628 & 0.483 & 0.378 & 0.302 & 0.238 & 0.518 & 0.889\\
   \multicolumn{1}{l}{JMAN (DH, rgb)} & 0.636 & 0.484 & 0.390 & 0.312 & 0.234 & 0.517 & 0.882\\
   \multicolumn{1}{l}{JMAN (DH, flow)} & 0.641 & 0.493 & 0.392 & 0.306 & 0.233 & 0.520 & 0.895 \\
   \multicolumn{1}{l}{JMAN (DH, aud)} & 0.626 & 0.479 & 0.372 & 0.294 & 0.230 & 0.511 & 0.845 \\

   \hline  
   \multicolumn{1}{l}{JMAN (DH, C, S)} & 0.644 & 0.488 & 0.383 & 0.302 & 0.238 & 0.518 & 0.891\\
   \multicolumn{1}{l}{JMAN (DH, rgb, flow)} & 0.648 & 0.499 & 0.390 & 0.309 & \textbf{0.240} & 0.520 & 0.890 \\
   \multicolumn{1}{l}{JMAN (DH, C, S, rgb, flow)} & \textbf{0.657} & \textbf{0.510} & \textbf{0.400} & \textbf{0.318} & 0.238 & \textbf{0.527} & \textbf{0.911} \\
   \multicolumn{1}{l}{JMAN (DH, C, S, rgb, flow, aud)} & 0.641 & 0.496 & 0.390 & 0.312 & 0.234 & 0.516 & 0.882 \\
   \hline    
   
 \hline
 \multicolumn{8}{c}{multiple reasoning step}\\
 \hline
 \multicolumn{1}{l}{JMAN (DH, C, S, rgb, flow, \(n\) = 2)} & 0.658 & 0.513 & 0.406 & 0.325 & 0.239 & 0.523 & 0.917 \\
 \multicolumn{1}{l}{JMAN (DH, C, S, rgb, flow, \(n\) = 3)} & 0.663 & 0.517 & 0.408 & 0.327 & 0.239 & 0.527 & 0.917 \\
 \multicolumn{1}{l}{JMAN (DH, C, S, rgb, flow, \(n\) = 4)} & 0.662 & 0.517 & 0.412 & 0.333 & \textbf{0.242} & 0.532 & 0.935 \\
 \multicolumn{1}{l}{JMAN (DH, C, S, rgb, flow, \(n\) = 5)} & \textbf{0.667} & \textbf{0.521} & \textbf{0.413} & \textbf{0.334} & 0.239 & \textbf{0.533} & \textbf{0.941} \\
 
 \hline
\end{tabular}}
\end{center}
\caption{ The objective evaluation values of each model using the DSTC7-AVSD test set. The first part is the performance  of the baseline model proposed by DSTC-AVSD organizers. The second part and the third part show the objective evaluation values of proposed JMAN with 1 reasoning step (\(n\) = 1). The second part simplest modality to evaluate the effectiveness of each features. In the third part, we estimate the performance of the combination of different modalities, which are audio, vision, and language. Considering only textual modality and visual modality, the fourth part show the results for proposed JMAN with increasing reasoning step \(n\). The word in the parentheses means the given feature. (DH: dialogue history; C: video caption; S: video summary; rgb: i3d-rgb feature; flow: i3d-flow feature; aud: audio vggish feature)}
\label{tb:mono}
\end{table*}

From the second part of Table \ref{tb:mono}, we find that models using visual features can produce more accurate answers than models using textual features. To be more specific, all evaluation metrics of JMAN(DH, rgb) and JMAN(DH, flow) outperform JMAN(DH, C) and JMAN(DH, S). As the caption and the summary for each video in the AVSD dataset generally consist of two sentences, visual features are relatively more informative. However, we surprisingly find that the model with audio feature (JMAN(DH, aud)) performs worst among all models with the additional mono-type feature. We surmise that Vggish audio feature are noisier than textual and visual features.

After analyzing the models with additional mono-type feature, we then evaluate the performance of the model combining different features. With one reasoning step, JMAN(DH, C, S) in the third part of Table \ref{tb:mono}  take textual features (caption and summary) into account. To be more specific, the context vector \(z_1\) of JMAN(DH, C, S) is the concatenation of \(C_1\) and \(S_1\). Likewise, JMAN(DH, rgb, flow) considers visual features (RGB and optical flow) in first reasoning step, and the context vector \(z_1\) of this model is the concatenation of \(R_1\) and \(F_1\). The results show that the models combining two features (JMAN(DH, C, S) and JMAN(DH, rgb, flow)) have a better performance than the models with additional mono-type feature. Examining textual domain, JMAN(DH, C, S) slightly outperforms both JMAN(DH, C) and JMAN(DH, S). Moreover, JMAN(DH, rgb, flow) surpasses both JMAN(DH, rgb) and JMAN(DH, flow) for visual domain. We observe that the model combining visual features (JMAN(DH, rgb, flow)) exhibit better performance than the model combining textual features (JMAN(DH, C, S)). Similar to the results of models with additional mono-type feature, we think that visual features will help our system to generate better responses.

In order to fully comprehend videos, we then take the advantage from both visual and textual domain. Therefore, JMAN(DH, C, S, rgb, flow) unitizes both visual features and textual features and the context vector \(z_1\) of this model is the concatenation of \(R_1\), \(F_1\), \(C_1\), and \(S_1\) in the first reasoning step. Taking both visual features and textual features, all evaluation metrics of JMAN(DH, C, S, rgb, flow) are higher than JMAN(DH, C, S) and  JMAN(DH, rgb, flow). This result shows that the model can improve video understanding when effective information increases. Moreover, the improvement of the JMAN(DH, C, S, rgb, flow) model confirms the usefulness of visual and textual features provided by AVSD dataset. However, we found that adding audio feature to JMAN(DH, C, S, rgb, flow) deteriorates the performance. Because of the decreasing performance of JMAN(DH, C, S, rgb, flow, aud), we did not use audio feature to build our model when the reasoning step increases.

\begin{table*}
\begin{center}
\small
\scalebox{1.04}{
\begin{tabular}{ccccccccc}
\hline
   & BLEU-1  & BLEU-2  & BLEU-3  & BLEU-4 & METEOR & ROUGE-L & CIDEr & Human  \\
 \hline
  \multicolumn{1}{l}{Baseline ~\cite{hori2018end}}& 0.614 & 0.467 &	0.365 &	0.289 &	0.21 & 0.48	& 0.651	& 2.885      \\

 \hline
   \multicolumn{1}{l}{JMAN (DH, C, S, rgb, flow, \(n\) = 5)} &  \textbf{0.645} & \textbf{0.504} & \textbf{0.402} & \textbf{0.324} & \textbf{0.232} & \textbf{0.521} & \textbf{0.875} & \textbf{3.123}  \\
   						
 \hline
\end{tabular}}
\end{center}
\caption{Released by the AVSD organizers, this table shows the final result of objective evaluation values and human rating by using the DSTC8-AVSD test set.}
\label{tb:off}
\end{table*}

\subsection{Multi-step Reasoning}
From previous experimental results, we find that using attention mechanism on both visual and textual features improves the performance of video understanding. We further evaluate the video understanding performance of the proposed JMAN for different reasoning steps, leveraging both textual and visual features, i.e., the current question, the dialogue history, the caption, the summary, RGB, and spatial flow of videos. After the first reasoning step (\(n =1\)), JMAN then focuses on specific regions of the textual representation and visual representation that correspond to the input question. To identify the salient regions form the multi-modal features, we designed \(V_n\) and \(T_n\), which are aggregated from the uni-modal attended features after first reasoning step. For instance, the joint-attended textual feature \(T_n\) is generated by aggregating the attended caption feature \(C_n\) and the attended summary feature \(S_n\).

Comparing JMAN(DH, C, S, rgb, flow) to JMAN(DH, C, S, rgb, flow, \(n=2\)) in Table \ref{tb:mono}, merely increasing a single reasoning step to two improves performance. This result shows that adding important information from a cross-modal way and adding reasoning step help the model better understand videos and then be able to generate correct answers. Moreover, the results also show that the accuracy of JMAN consistently increases when reasoning step \(n\) grows. This advantage may come from the additional cross-modal joint-attended features (\(T_n\) and \(V_n\)) which bring in more information to the model. Nevertheless, for reasoning steps n beyond 5, the model did not show significant increase on every metrics. The best performance of our model (JMAN(DH, C, S, rgb, flow, \(n=5\)) achieves 20.8\% improvement over the baseline on CIDEr score for DSTC7-AVSD dataset. Therefore, we submitted this best model to DSTC8-AVSD track. Table \ref{tb:off} is the final result released by the official. Our submitted system outperforms the released baseline model for both subjective and objective evaluation metrics. 

\begin{figure}[t]
\begin{center}
\includegraphics[scale=0.29, trim={0 0 0 0}]{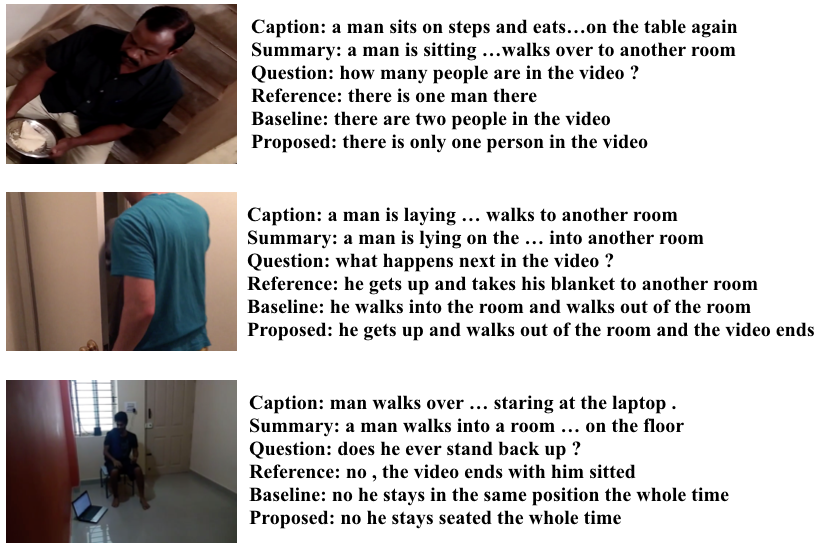}
\end{center}
   \caption{Examples of reference answers and the answers generated by the baseline model and the proposed JMAN model on DSTC7-AVSD dataset. Only parts of the video caption and the video summary are shown for simplicity. The pictures are the  frames from Charades video dataset used by DSTC7-AVSD dataset.}
\label{fig:example}
\end{figure}

\subsection{Qualitative Analysis and Training Data Quality}

Figure \ref{fig:example} shows the ground truth reference proposed by the AVSD dataset and the answers generated by the baseline model and the proposed JMAN model. The generated answers illustrate that multiple reasoning steps benefit the inference process and hence lead to accurate answers of questions. For example, the proposed model can focus on the people in the frame and correctly answer the number of people in the dynamic scenes video. Compared with “walks into and walks out of the room”, the open-end question “what happens next in the video ?” is provided with a more detailed answer “gets up and walks out of the room and the video ends”. Moreover, we found that the proposed model can generate more precise answers according to complex features through the joint-modality attention network. Compared with “the same position” generated by the baseline model, the question “does he ever stand back up ?” is provided  with a more precise answer “he stays seated the whole time” by the proposed model.

We observe some issues that might affect the performance of video understanding in AVSD dataset. Some ground-truth answers provided an ambiguous answer that could lead the model hard to learn. For example, the question “what does this room appear to be ?” is answered with “hard to say”. Moreover, the reference sometimes gives answers beyond the question. For example, for the question “does she just hold the towel ?", the ground-truth answer is “yes , she hold it , smile and spoke a few words of spanish" which “smile and spoke ..." is beyond the question. Furthermore, many to-be-answered questions in the training data ask for additional information, such as “anything else that i should know ?” is answered with “no that is it in the video”. Therefore, more precise question-and-answer pairs would benefit model learning.

\section{Conclusion}

This paper proposes an encoder-decoder based visual dialogue model which consider multiple modalities effectively by the proposed joint-modality attention network (JMAN). Jointly taking both visual features and textual features at each reasoning step, JMAN extracted important part from cross-modal features and achieved a better comprehension of multi-modal context. Through multiple reasoning steps, our model further boosted the performance of scene-aware ability. Our best model achieved a significant 12.1\% and 22.4\% improvement over the baseline on ROUGE-L and CIDEr. We hope to explore this multi-modal dialogue setting further in the future with larger scale datasets. Unsupervised pre-trained language model could also applied to inject more semantics to the model for multi-modal dialogue task.

\section{Acknowledgement}
This research is supported by the Ministry of Science and Technology, Taiwan under the project contract 108-2221-E-001-012-MY3.


\vspace{-0.5pc}
\bibliography{bibliography}

\bibliographystyle{aaai}
\end{document}